\documentclass[conference]{IEEEtran}
\IEEEoverridecommandlockouts
\usepackage{cite}
\usepackage{amsmath,amssymb,amsfonts}
\usepackage{algorithmic}
\usepackage{graphicx}
\usepackage{textcomp}
\usepackage{xcolor}
\usepackage{multirow}
\usepackage{caption}
\usepackage{subfig}
\def\BibTeX{{\rm B\kern-.05em{\sc i\kern-.025em b}\kern-.08em
    T\kern-.1667em\lower.7ex\hbox{E}\kern-.125emX}}
\begin{document}

\title{Application of the Neural Network Dependability Kit in Real-World Environments\\
\thanks{This publication is part of a project that has received funding from the European Union's Horizon 2020 research and innovation programme under grant agreement No. 761708.}
}

\author{\IEEEauthorblockN{Amit Sahu}
\IEEEauthorblockA{\textit{fortiss GmbH} \\
Munich, Germany \\
sahu@fortiss.org}
\and
\IEEEauthorblockN{Noelia V\'{a}llez}
\IEEEauthorblockA{\textit{Ubotica Technologies}\\
Dublin, Ireland \\
noelia.vallez@ubotica.com}
\and
\IEEEauthorblockN{Rosana Rodr\'{i}guez-Bobada}
\IEEEauthorblockA{\textit{Ubotica Technologies}\\
Dublin, Ireland \\
rosana.rodriguez@ubotica.com}
\and 
\IEEEauthorblockN{Mohamad Alhaddad}
\IEEEauthorblockA{\textit{ISSD Bili\c{s}im Elektronik A.\c{S}.} \\
Ankara, Turkey \\
muhammed@issd.com.tr}
\and
\IEEEauthorblockN{Omar Moured}
\IEEEauthorblockA{\textit{ISSD Bili\c{s}im Elektronik A.\c{S}.} \\
Ankara, Turkey \\
omar@issd.com.tr}
\and
\IEEEauthorblockN{Georg Neugschwandtner}
\IEEEauthorblockA{\textit{fortiss GmbH} \\
Munich, Germany \\
neugschwandtner@fortiss.org}
}

\maketitle

\begin{abstract}
In this paper, we provide a guideline for using the Neural Network Dependability Kit (NNDK) during the development process of NN models, and show how the algorithm is applied in two image classification use cases. 
The case studies demonstrate the usage of the dependability kit to obtain insights about the NN model and how they informed the development process of the neural network model. 
After interpreting neural networks via the different metrics available in the NNDK, the developers were able to increase the NNs' accuracy, trust the developed networks, and make them more robust. 
In addition, we obtained a novel application-oriented technique to provide supporting evidence for an NN's classification result to the user. 
In the medical image classification use case, it was used to retrieve case images from the training dataset that were similar to the current patient's image and could therefore act as a support for the NN model's decision and aid doctors in interpreting the results. 
\end{abstract}


\section{Introduction}

Neural networks have obtained state of the art results in vision-based perception (e.g. YOLO~\cite{yolov4}). 
This makes them essential for applications like autonomous driving, medical image processing, etc. 
However, the safety-critical nature of these applications limits the applicability of neural networks in the real world due to their black box nature (weights and parameters). 
At fortiss, we tackled this challenge by developing a neural network dependability kit (NNDK)~\cite{nndk}. 
NNDK offers dependability metrics for developers to observe the learning (training) process, formal reasoning to avoid risky behaviour (risk properties), and runtime monitoring to indicate the network's application to unknown examples. 

Theoretical analysis of NNDK and its application on standard datasets had been discussed in previous research papers~\cite{dep_metrics, max_res, runtime_monitor}. 
In this paper, we discuss two use cases of applying NNDK during the development phase in real-world environments:
\begin{itemize}
 \item Medical Imaging: Diabetic Retinopathy Detection by Ubotica
 \item Smart Tunnels: Incident Detection by ISSD
\end{itemize} 
fortiss worked with the development teams at Ubotica, and ISSD and supported the application of the NNDK in the development process as part of the FED4SAE project. 

NNDK offers support (dependability metrics~\cite{dep_metrics}) during different phases of the product life cycle:
\begin{enumerate}
 \item Data preparation: scenario k-projection coverage
 \item Training and validation: neuron activation pattern (NAP) metric, risk properties using formal reasoning
 \item Testing and generalization: neuron k-projection coverage, perturbation loss metric
 \item Operation: runtime monitoring using NAP
\end{enumerate}

NNDK techniques were applied during these phases and based on the results, the next steps were recommended. After following up on these recommendations, better outcomes in terms of the dependability metrics were obtained. 
The outcomes influenced the development decisions and the developers were able to achieve more efficient (faster prediction and low memory usage) and effective (higher accuracy) operation of the neural network.   

\section{Development process with NNDK techniques}
\label{sec:dev_proc}

\begin{figure*}[htbp]
\centerline{\includegraphics[width=\textwidth]{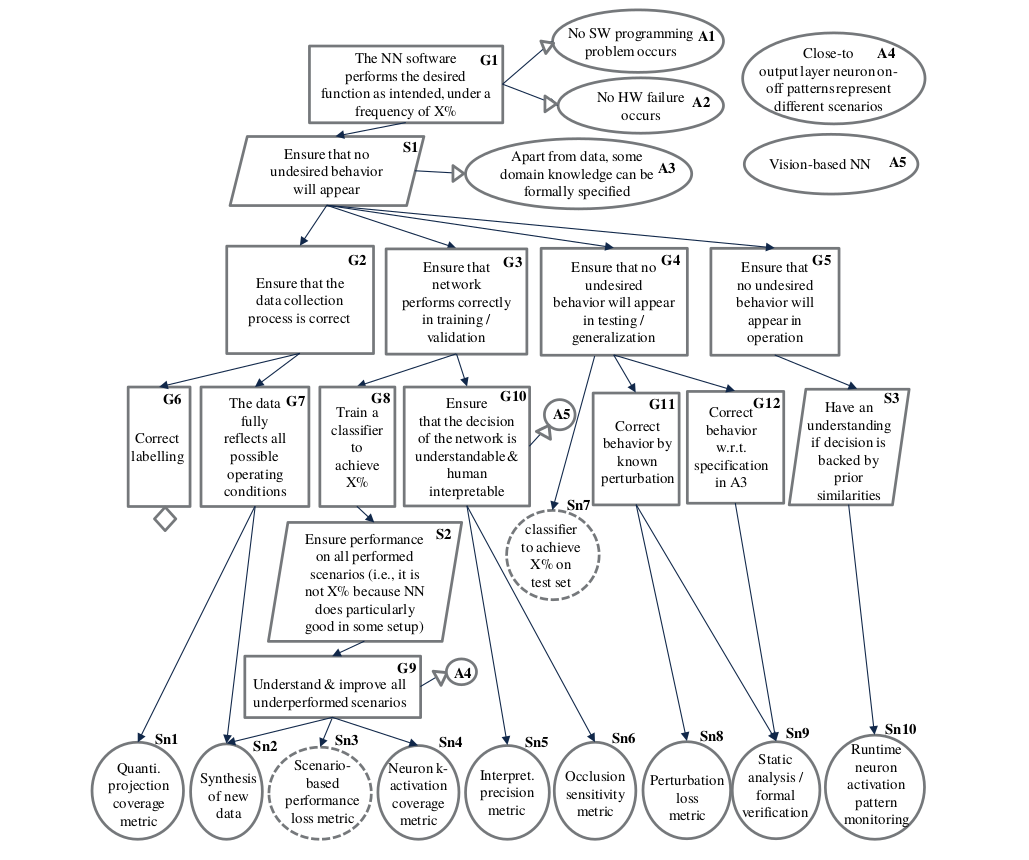}}
\caption{A simple GSN based interface to select the relevant solutions (Sn) from NNDK based on the required goals (G), including assumptions (A) and the agreed strategies (S) \cite{nndk}. }
\label{fig:gsn}
\end{figure*}

As stated in the previous section, NNDK techniques can be applied during different phases of the product life cycle. 
In the following, we describe the generic development phases that are commonly used by ML development teams along with the NNDK metrics that will allow these teams to quantify and improve the quality of the process in the respective phases.

\textbf{Data Preparation:} Data collection is the most basic and essential task as every other phase is affected by its quality. 
High-quality datasets cover all scenarios and situations. However, obtaining a complete set of scenarios can result in a combinatorial explosion. 
Hence, to quantify the coverage of the scenarios by the dataset, NNDK offers scenario k-projection coverage.
It is based on the concept of combinatorial testing~\cite{combinatorial} and provides a relative form of completeness against the combinatorial explosion of scenarios~\cite{coverage}. It can also suggest new scenarios for labeling that will maximally improve the dataset. 

\textbf{Training and Validation:} NN model training methods already involve metrics, in line with the application tasks, which widely differ based on their requirements. Hence, NNDK metrics were more directed towards validation and avoiding risky behaviour. For example, formal reasoning could ensure that the model satisfies the risk properties to ensure predictable/reliable behaviour under conditions that are similar, but possibly different, to the ones experienced in the test cases. Due to computational limits, the tool provides a layer-wise analysis of the NN model. Since most NN architectures have deep layers, it provides analysis over a selected shallow layer and can check the possibility of erroneous behaviour on the boxed domain of its neurons. 
In the case of erroneous behaviour further techniques (such as Counter Example Guided Abstraction Refinement, CEGAR) can be used to identify counter-examples that can help in training a better model.

In addition, one can look into the histograms of the Neuron Activation Pattern (NAP) of each class. One starts by selecting a target layer and obtaining its on-off activation pattern. As the training progresses, data points of the same class should have a similar number of neuron (feature) activations. Hence, the histogram of a good model is a narrow graph. Observing the histogram as the training progresses lets one keep this property in check. 

\textbf{Testing and generalization:} Test cases should cover all paths of the software and be robust to noisy environments. In the case of NNs, these paths correspond to neuron activations. However, checking all possible neuron activations can result in a combinatorial explosion. Therefore, in addition to scenario k-projection coverage, NNDK also offers neuron k-projection coverage over a pre-selected layer. This measures the completeness of the test set to cover the whole neuron layer under analysis. 

Also, since real environments are prone to different kinds of noise based on hardware (Gaussian), weather (snow, haze), or even deliberate (adversarial noise), NNDK offers a perturbation loss metric to measure how the system performs in these situations. The metric creates these noisy data points and tests the model's robustness (generalization) to noisy environments. These noisy data points can further be used for dataset augmentation. Training with an augmented dataset increases the model's robustness and can be measured using the perturbation loss metric. 

\textbf{Operation:} The behaviour of the NN model during training and in the operation environment can be widely different if the two datasets do not have an equivalent data distribution. Hence, we can only expect adequate performance from the NN model when it is applied to a data point with prior similarities to the training data. NNDK keeps track of data points by recording their Neural Activation Pattern (NAP) on a pre-selected (mostly the penultimate) layer. During runtime, the NAP of the data point (in operation) is compared with the NAP database of the training data using the hamming distance. If the pattern is not found in the database, a warning is generated that the model's output is not supported by the training data.

As a guideline for applying NNDK as an ad-hoc solution limited to certain dependability aspects of the development process, 
the Goal Structured Notation (GSN) diagram shown in Figure~\ref{fig:gsn} shows the contribution of NNDK metrics towards the overall goal~\cite{nndk} of a safe system. In this diagram, we tried to cover many safety situations that will be further extended as we progress with the development of NNDK. The figure acts as an aid to select the solution (Sn) provided by NNDK based on the associated goals (G), assumptions (A), and strategies (S). For example, one can start from the top goal G1 of having an intended function delivered by the NN model. One strategy S1 would be to ensure that no undesired behaviour occurs. This can be refined into the different phases of the development cycle. During the operation phase (goal G5) strategy S3 would be comparing the decision with prior similarities. Using this final strategy NNDK provides solutions in terms of runtime monitoring (Sn10).

\section{Success stories of NNDK application}
\label{sec:success_stories}
The case studies in this paper were made in 2019/2020 as part of the FED4SAE project.
NNDK was applied at all stages to generate reports, which were used for guiding the NN model development. 

Due to the close collaboration between fortiss and the development teams, the simplified process from Figure~\ref{fig:gsn}  was not used. Instead, the metrics were selected following an in-depth  discussion and analysis of the respective requirements and development processes.

\subsection{Medical Imaging: Diabetic Retinopathy}
The objective of the development was to create a working prototype that demonstrates the classification of retinal fundus images for the presence of Diabetic Retinopathy (DR) indicators. Some of these indicators are tiny abnormal leaky blood vessels (neovascularization), pale fatty deposits on the retina (exudates), and damaged nerves fibers ("cotton wool spots").

The initial model was InceptionV3, whose last layer was changed using transfer learning to classify the fundus images. The last layer consisted of 2048 neurons (features) which were mapped using a fully connected layer to two classes. Since NNDK requires the selection of a feature layer to quantify the learning process and the last layer (of the model) included only features obtained using training on the ImageNet dataset, new layers were added to the model. Two fully connected extra layers with size 1024 and 512 were added and trained on Kaggle’s Diabetic Retinopathy Detection challenge dataset.

We selected relevant metrics by considering each development phase sequentially:
\begin{enumerate}
 \item Since the dataset consisted of only images with few labels, the scenario coverage metric for data preparation was not considered.
 \item Due to limited understanding of the features obtained, formal reasoning was not applied during the training and the validation phase as it requires the creation of risk properties. The Neuron Activation Pattern (NAP) metric for both classes (positive, negative) was applied and visualized using the histogram graph.
 \item For testing and generalization phases, both neuron k-projection coverage and the perturbation loss metric were applied.
 \item In the operation phase, runtime monitoring was employed to warn if the model's output is not supported by the training data.
\end{enumerate}

The tables and figures that explain the results of these metrics are attached in Appendix~\ref{app:DR} for more details.
Here, we only explain important points and the insights from the above metrics that were useful in the development process.

Important points and insights from the NAP metric histogram for both classes--positive, negative--are as follows: 
\begin{itemize}
 \item The penultimate layer (512 neurons) was selected for analysis as it was assumed to have the most high-level features.
 \item The graph for both classes peaked at high neuron activations (around 2000 from 2048) for the initial model.
 \item As the new model was trained, the graph showed reductions in the number of activations for the negative cases and promotion of activations for the positive cases.
 \item Since identifying the presence of a disease (positive class) should be more feature-oriented than the absence of the disease (negative class), the training process was more trustworthy.
\end{itemize}

Important points and insights from the neuron k-projection coverage:
\begin{itemize}
 \item In the initial model, the coverage was high over the dataset as the ImageNet-trained model considered all the features before classifying.
 \item In the new model, as the training progressed, the coverage increased first and then kept going downwards.
 \item The increment was attributed to the learning of specific features from the dataset and the decrement was attributed to the unlearning of features from the ImageNet dataset.
 \item In the final model, the reduced coverage suggested that even with a smaller set of neurons in the layer, the classification will most likely remain the same. Hence, in later stages of development, pruning of the network architecture would be useful. After 74\% pruning, the model still achieved similar accuracy (reduction by 0.1) with less than half of the prediction time as compared to the non-pruned model. 
\end{itemize}

Important points and insights from the perturbation loss metric:
\begin{itemize}
 \item After applying different kinds of noise--s\&p, Poisson, blur, brightness, gain, Gauss--on the final model; Gauss and s\&p were found to be most effective.
 \item Positive data points had an average loss in confidence score of around 50\% from Gaussian noise but negative data points only suffered a 10\% loss.
 \item Negative data points had an average loss of around 80\% from s\&p noise but positive data points only suffered a 5\% loss.
 \item Developers speculated that the effect of Gaussian noise on positive data points was due to hiding the definition of retinal lesions. This reduced the number of detected features and therefore the classification.
 \item Developers speculated that the effect of s\&p noise on negative data points was due to misunderstanding of the small noise points as small lesions of diabetic retinopathy.
 \item The unaffected  class cases support their speculation as well.  
\end{itemize}

Important points and insights from runtime monitoring:
\begin{itemize}
 \item Since the dataset was small and unbalanced (9,316 positive cases and 25,810 negative cases) there was a high possibility that a fully comprehensive model including all types of cases would be trained.
 \item In order to differentiate from the cases that were not part of the training, runtime monitoring using the NAP metric was employed.
 \item As expected, there were a few cases that were not found in the patterns from the training dataset. 
 \item These non-priors were classified with high confidence ($>= 92\%$). This shows the limitation of neural networks in identifying unknown cases using the confidence score.
 \item Furthermore, since there were no samples with a mismatch between the label predicted and the label of the closest pattern match, the model effectively assigned each class according to the patterns learnt during training.
\end{itemize}

In addition to applying NNDK for dependability metrics, we discovered a novel application to use these metrics for case-based reasoning. NNs only provide a confidence score, which (as seen in the runtime monitoring case) can be deceptive for unknown cases. Hence, in addition to providing the confidence score, during operation, a set of similar images from the training dataset would be beneficial. In the case of DR, Ubotica used this feature in their application to provide doctors with similar case images using the NAP. Neural networks first try to extract features in the NN layer and then use this layer for classification. One can also use the visualization of these features on the images from the training dataset to serve as evidence for the NN model's decision. A doctor can analyze these supporting images to agree or disagree with the model's decision.

\subsection{Object Detection: Smart Tunnels}
The objective of the development was an Automatic Incident Detection (AID) system for road tunnels using neural networks. In improvement to the current OpenCV \cite{opencv} based solution, it was anticipated that the final system would be able to both detect and track the movement of the various vehicles or pedestrians with higher accuracy. The development team was able to achieve pedestrian detection and stationary vehicle detection in the final application. The application was validated in an operational tunnel where it performed better than the legacy system.

A standard object detection model--YOLOv3\cite{yolov3}--was used for this task. A transfer learning model from YOLOv3 for the two classes was evaluated but the improvements were not significant. 

Following the process according to Section~\ref{sec:dev_proc}, relevant metrics were selected by considering each development phase sequentially:
\begin{enumerate}
 \item A tunnel is a very controlled and constant environment. For example, except close to the entrance and exit, the lighting and precipitation never changes. Thus, the diversity in scenarios is minimal and coverage can be controlled manually. Hence, the scenario coverage metric for data preparation was not considered.
 \item Due to limited understanding of the features obtained and the use of a standard NN model (YOLOv3), formal reasoning was not applied during the training and the validation phase as it requires the creation of risk properties specific to the use case. The neuron activation pattern (NAP) metric for both classes (positive, negative) was applied and visualized using the histogram graph for validating the training results.
 \item For testing and generalization phases, both Neuron k-projection coverage and the perturbation loss metric were applied.
 \item During the operation environment, runtime monitoring was not deemed relevant due to the presence of cross-checking by human tunnel operators. 
\end{enumerate}

In this use case, important points and insights from the dependability metrics were centered on network pruning and noise sensitivity.

 Neuron k-projection coverage and NAP metrics showed that very few neurons were activated for both classes (pedestrian, stationary vehicle). This suggested that network pruning would be a good next step.
 This was an essential insight as efficiency of the inference step was a key requirement in this use case due to the high number of images that need to be processed continuously.
 
Applying the perturbation loss metric, it was discovered that the images are highly prone to the noise, resulting in a 95\% average loss of confidence.
To deal with this problem, dataset augmentation was done and new models were trained.

Two models emerged from such training, one was more robust to weather-related noise, and the other was more robust to hardware related noise.
Weather-related noise like snow or haze can occur in some cameras at the entry and exit.
Hardware related noise like Gaussian (due to high temperature, low illumination), salt \& pepper (due to sharp disturbance in signal), and Poisson (due to electromagnetic properties of light particles) can still occur inside the tunnel.
Therefore, as hardware noise affects more cameras in a tunnel, the final selected model was the one that was more robust against it.

\subsection{Conclusion}
We started with a brief explanation of the role that NNDK can play in the development of a neural network model, followed by an explanation of the usage of the metrics in the context of different development phases. We explained the results and use cases of the NNDK looking at two real-world applications: classification of medical images for disease markers and detecting pedestrians and stationary vehicles on road tunnel surveillance camera feeds. Additionally, we introduced a novel technique to report similar data points from prior similarities in the training dataset using the Neural Activation Pattern (NAP) metric. This new technique provides support for the NN model's decision and helps with interpreting the results. As seen from the results on real-world applications, NNDK helped in increasing the accuracy, developing trust in the models, and making them more robust.   

\section*{Acknowledgment}
The authors are grateful for the valuable insight gained from the discussions with Finian Rogers of Intel R\&D Ireland throughout the project.

\begin{appendices}

\section{Medical Imaging: Diabetic Retinopathy}
\label{app:DR}

Most people with diabetes are unaware of having diabetes complications. However, most complications can be detected in their early stages by screening programs. In particular, Diabetic Retinopathy (DR) is a medical condition derived from diabetes in which damage occurs to the retina. Individuals who present for screening have retinal fundus images taken using a specialist camera. These images are read by screeners at a later date to detect the presence of DR indicators. 

An InceptionV3 was initially trained to distinguish between DR and NonDR images. The datased used was the one published in the Kaggle's Diabetic Retinopathy Detection challenge and the training was based on a transfer learning approach where the ImageNet weights were loaded and only the dense layers were trained.

The architecture was then modified adding two dense layers of size 1024 and 512 to better facilitate the analysis. In this case, the complete architecture was trained. The model selected was the one corresponding to the checkpoint from epoch 110.

Finally, the new model undertook a pruning process to reduce their size while roughly maintaining the accuracy. The approach followed consists of simultaneously deleting a 2\% of the network channels in a loop. The channels to be removed each time are the channels which have the highest Average Percentage of Zeros (APoZ). In addition, to allow the model to compensate for the pruned channels, 10 training epochs are run between pruning iterations.

\begin{figure*}[!htbp]
	\centering
	\subfloat[][initial model]{
		\includegraphics[width=0.4\linewidth]{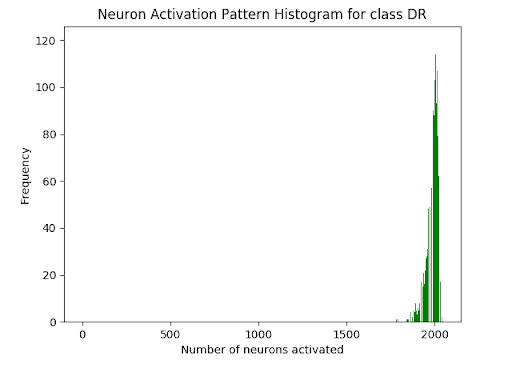}
		\includegraphics[width=0.4\linewidth]{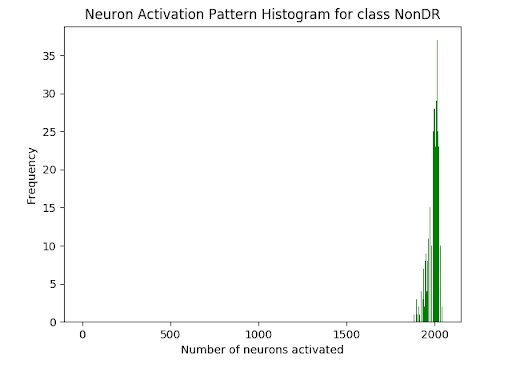}
	}\\
	\subfloat[][new model (epoch = 1)]{
		\includegraphics[width=0.4\linewidth]{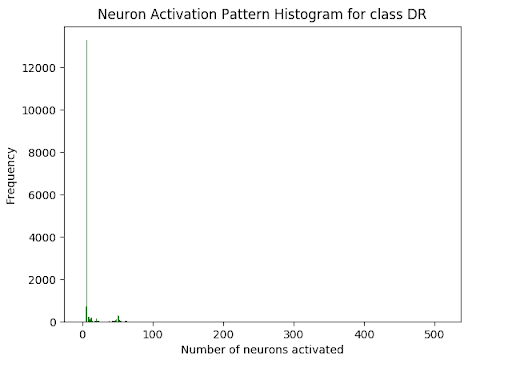}
		\includegraphics[width=0.4\linewidth]{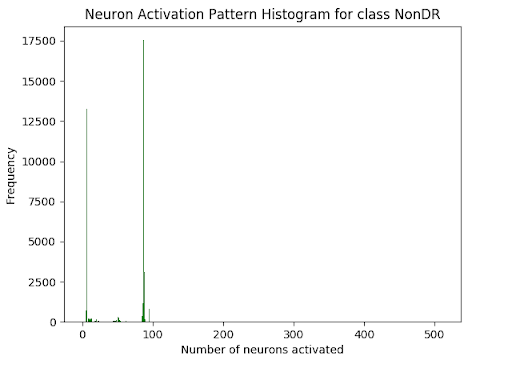}
	}\\
	\subfloat[][new model (epoch = 60)]{
		\includegraphics[width=0.4\linewidth]{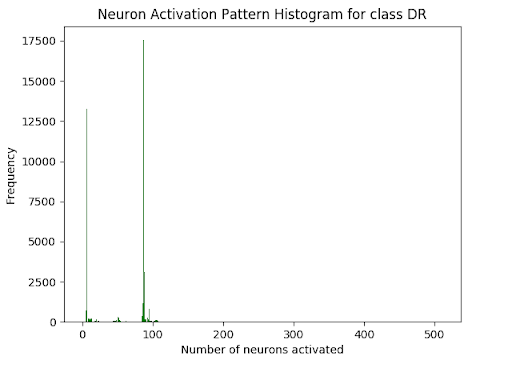}
		\includegraphics[width=0.4\linewidth]{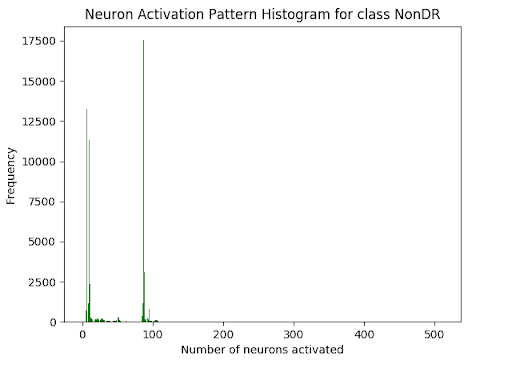}
	}\\
	\subfloat[][new model (epoch = 110)]{
		\includegraphics[width=0.4\linewidth]{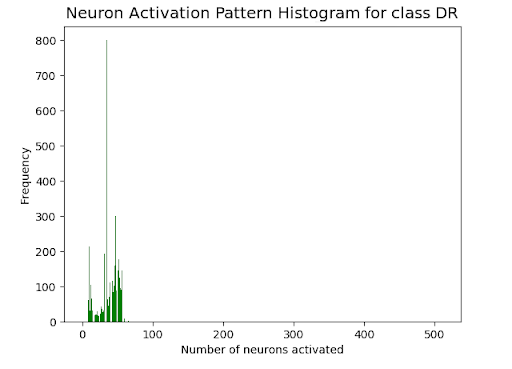}
		\includegraphics[width=0.4\linewidth]{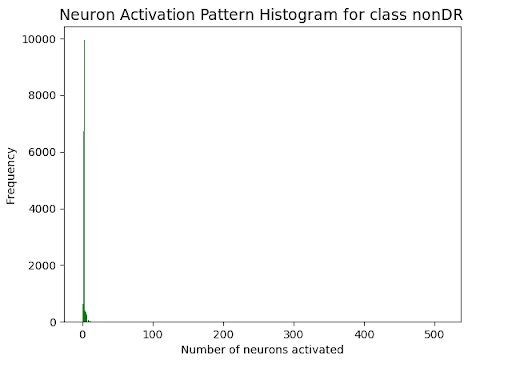}
	}\\
	\caption{NAP metric results}
	\label{ubotica_nap}
\end{figure*}

\begin{figure}[!htb]
\centering
\includegraphics[width=0.8\linewidth]{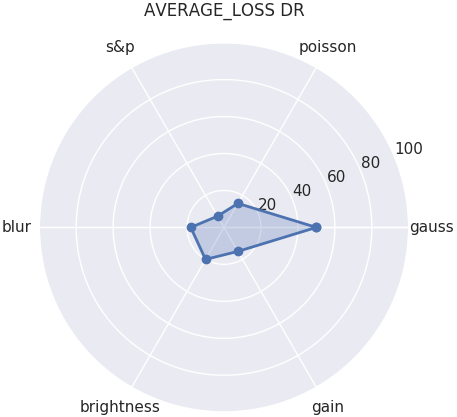} \\ \vspace{0.3cm}
\includegraphics[width=0.8\linewidth]{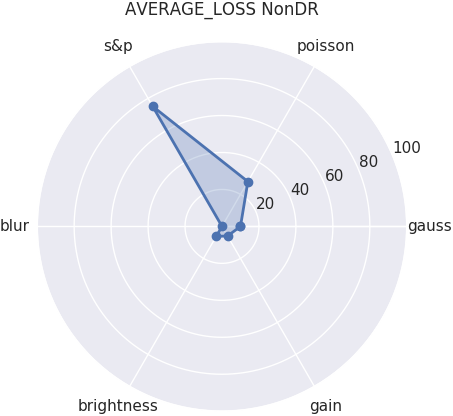}
\caption{Perturbation loss results of the new model (epoch = 110)}
\label{ubotica_perturbation}
\end{figure}

\begin{figure*}[!htb]
\centerline{\includegraphics[width=\linewidth]{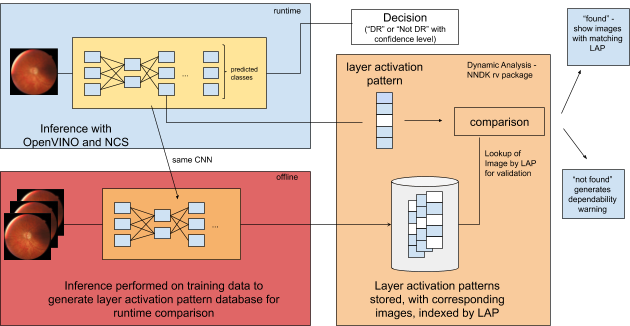}}
\caption{Runtime monitoring process}
\label{ubotica_dynamic}
\end{figure*}

Tables \ref{ubotica_kprojection1} and \ref{ubotica_kprojection2} show the k-projection coverage results of the initial and new models including some intermediate results. Similarly, Figure \ref{ubotica_nap} depicts the NAP metric results. The perturbation loss results of both the initial and the new model are shown in Figure \ref{ubotica_perturbation}. Finally, Figure \ref{ubotica_dynamic} and Table \ref{ubotica_dynamict} show how runtime monitoring is used in this particular scenario and the number of images that have a certain real\_class-predicted\_class-closest\_pattern\_class combination respectively.



\begin{table}[!htb]
\caption {Results of k-projection coverage with k=1 and k=2 from the initial model}
 \centering
 \renewcommand{\arraystretch}{1.2}
\begin{tabular}{|c|c|c|c|} 
\hline
\multicolumn{2}{|c|}{1-projection}  & \multicolumn{2}{c|}{2-projection} \\
\hline
DR  & NonDR  & DR  & NonDR  \\ 
\hline
3976 / 4096 & 3959 / 4096 & 7034545 / 8384512 & 6822593 / 8384512 \\ 
0.970703    & 0.966552    & 0.838992          & 0.813713\\ 
\hline
\end{tabular}
\label{ubotica_kprojection1} 
\end{table}

\begin{table}[!htb]
\caption {Results of k-projection coverage with k=1 and k=2 from the new model and 5 of its checkpoints}
\renewcommand{\arraystretch}{1.2}
 \centering
  \resizebox{\columnwidth}{!}{  
\begin{tabular}{|c|c|c|c|c|}
\hline
\multirow{2}{*}{Model} & \multicolumn{2}{c|}{1-projection} & \multicolumn{2}{c|}{2-projection}\\
\cline{2-5} 
& DR  & NonDR & DR & NonDR
\\ \hline
New model	& 625 / 1024 & 627 / 1024 & 188987 / 523264 & 190487 / 523264 \\
epoch=01	& 0.610351 	 & 0.612304	  & 0.361169 	   	& 0.364036 \\ 
\hline
New model   & 856 / 1024 & 827 / 1024 & 336990 / 523264 & 307938 / 523264 \\ 
epoch=20    &  0.835937  & 0.807617   & 0.644015        & 0.588494 \\ 
\hline
New model   & 787 / 1024 & 781 / 1024 & 288822 / 523264 & 281457 / 523264 \\ 
epoch=40    & 0.768554   & 0.762695   &  0.551962       & 0.537887 \\ 
\hline
New model   & 735 / 1024 & 718 / 1024 & 257012 / 523264 & 247240 / 523264 \\ 
epoch=60    & 0.717773   & 0.761171   & 0.491170        & 0.472495 \\ 
\hline
New model   & 607 / 1024 & 594 / 1024 & 181143 / 523264 & 173959 / 523264 \\ 
epoch=80    & 0.592773   & 0.580078   & 0.346178        & 0.332449 \\ 
\hline
New model   & 600 / 1024 & 610 / 1024 & 176847 / 523264 & 182512 / 523264 \\ 
epoch=110   & 0.58593    & 0.595703   & 0.595703        & 0.348795 \\ 
\hline
\end{tabular}
}
\label{ubotica_kprojection2} 
\end{table}


 \begin{table}[!htb]
\caption {Accuracy results of the pruned models. In the model name wXX-wYY, YY indicates the checkpoint from the training epochs between pruning iterations an XX the checkpoint from the training epochs after performing the pruning.}
\renewcommand{\arraystretch}{1.2}
 \centering
\begin{tabular}{|c|c|c|c|c|}
\hline
\multirow{2}{*}{\% Pruning} & \multirow{2}{*}{Pruned model} & \multicolumn{3}{c|}{Results}   \\ \cline{3-5} 
                            &                                  & DR Acc & NonDR Acc & Acc \\ \hline
0\%                         & new-model-w110                   & 0.630  & 0.830     & 0.770     \\ \hline
22\%                        & w07-w03                          & 0.602  & 0.845     & 0.781     \\ \hline
38\%                        & w25-w10                          & 0.640  & 0.795     & 0.755     \\ \hline
56\%       					& w07-w05                          & 0.579  & 0.844     & 0.775     \\ \hline 
74\%                        & w02-w10                          & 0.591  & 0.832     & 0.769     \\ \hline
\end{tabular}
\end{table}

 
 \begin{table}[!htb]
\caption {Distribution of the number of images using a Hamming distance of d=0 and according to the real class, the predicted class, and the class of the closest pattern found}
\renewcommand{\arraystretch}{1.2}
 \centering
\begin{tabular}{|l|l|l|c|c|}
\hline
\multicolumn{1}{|c|}{Real} & \multicolumn{1}{c|}{Predicted} & Closest Pattern & \# Images & Avg. Confidence\\ 
\hline
\multirow{8}{*}{DR}    & \multirow{4}{*}{DR}            & \textbf{DR}    & \textbf{534}  & \textbf{0.95} \\ 
					   &                                & NonDR          & 0             & -             \\ 
					   &                                & Both           & 28            & 0.53          \\ 
					   &                                & Not found      & 12            & 0.98          \\ \cline{2-5} 
					   & \multirow{4}{*}{NonDR}         & DR             & 0             & -             \\ 
					   &                                & NonDR          & 318           & 0.87          \\ 
					   &                                & Both           & 21            & 0.51          \\ 
					   &                                & Not found      & 2             & 0.92          \\ \hline
\multirow{8}{*}{NonDR} & \multirow{4}{*}{DR}            & DR             & 402           & 0.87          \\ 
					   &                                & NonDR          & 0             & -             \\ 
					   &                                & Both           & 34            & 0.52          \\ 
					   &                                & Not found      & 7             & 0.94          \\ \cline{2-5} 
					   & \multirow{4}{*}{NonDR}         & DR             & 0             & -             \\ 
					   &                                & \textbf{NonDR} & \textbf{2080} & \textbf{0.91} \\ 
					   &                                & Both           & 39            & 0.51          \\ 
					   &                                & Not found      & 6             & 0.96          \\ \hline
\end{tabular}
\label{ubotica_dynamict}
\end{table}


\clearpage

\section{Object Detection: Smart Tunnels}[htb]
\label{app:issd}
Thousands of vehicular tunnels are in constant use worldwide.  
As the main safety mechanism, these tunnels deploy CCTV cameras which are wired back to a central control room. 
Here, human operators monitor the vehicular flow to ensure the safety of the tunnel and its users. 
However, human operators alone cannot handle the monitoring task at large scales due to vigilance decrements. 
Hence, they are supplemented with automatic incident detection (AID) systems.

\begin{figure*}[!htb]
 \centerline{\includegraphics[width=\textwidth]{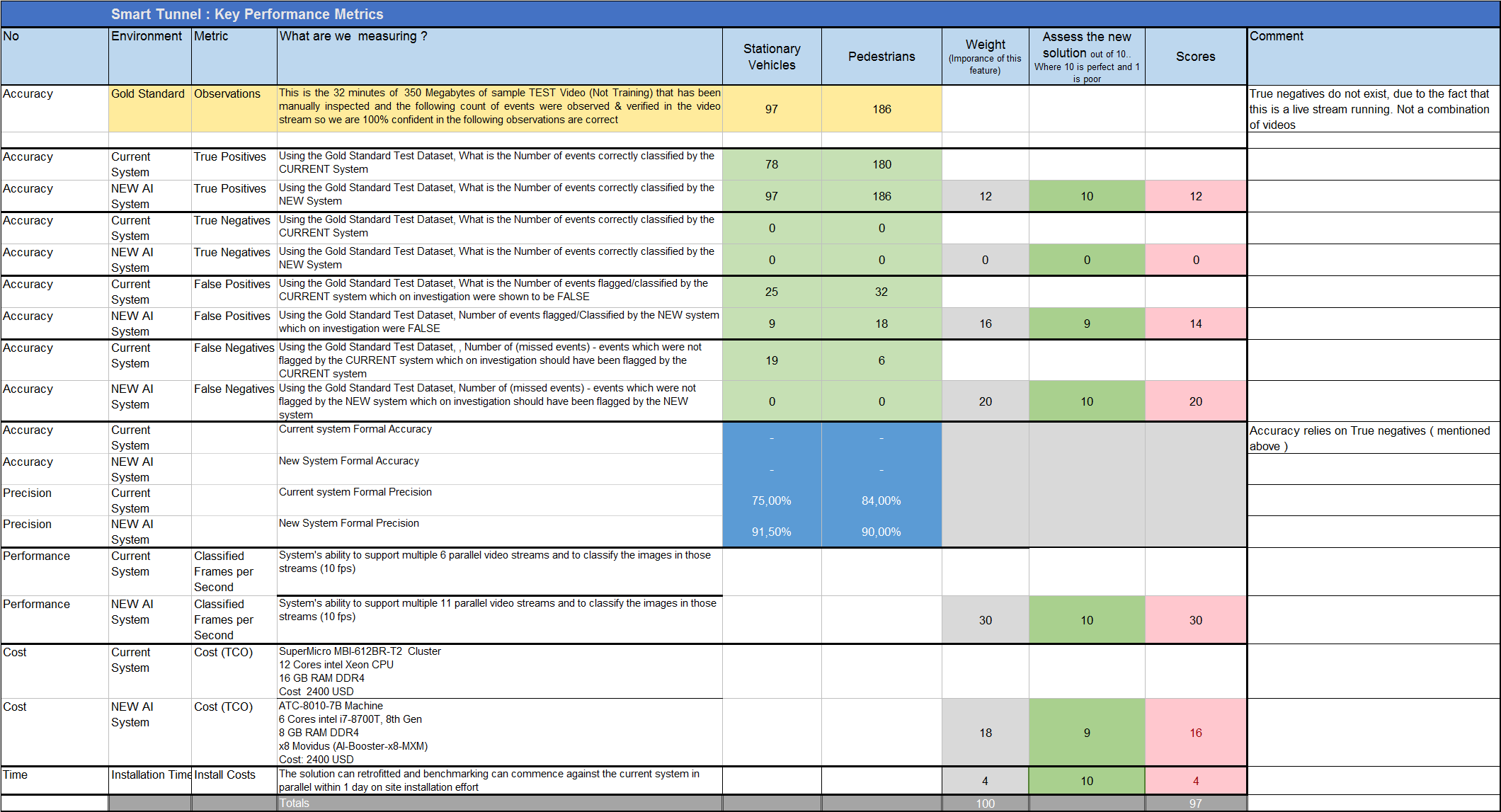}}
 \caption{Score card comparing key performance metrics of the two systems: SPECTO, Smart Tunnel.}
 \label{fig:score_card}
\end{figure*}

At ISSD, the previously deployed solution (SPECTO) employed algorithms from OpenCV. 
The project's challenge was to leverage its performance while delivering a reasonable reduction in the product life cycle cost. 
Hence, the AID solution was enhanced with object detection using neural networks--Smart Tunnel. 
The final system was able to detect the vehicles, and pedestrians with better performance and much more efficient hardware costs. 
The comparison of the two systems via scorecard is shown in Figure~\ref{fig:score_card}.

NNDK metrics were analysed in a structured way (Section~\ref{sec:dev_proc}) and the relevant ones were evaluated to understand the working of the NN model--YOLOv3. 
Neuron activation pattern metric was applied as follows:
\begin{enumerate}
 \item A convolution layer ($104^{th}$) with shape (255, 32, 52) was selected for observation.
 \item Average filter was applied over it to obtain a vector of size 255.
 \item Objects were filtered based on the confidence thresholds: 80\%, 95\%.
 \item For each class--vehicle, pedestrian--neuron activations of detected objects with the above confidence thresholds were noted. 
\end{enumerate}

\begin{figure}[!htb]
\centering
\includegraphics[width=\linewidth]{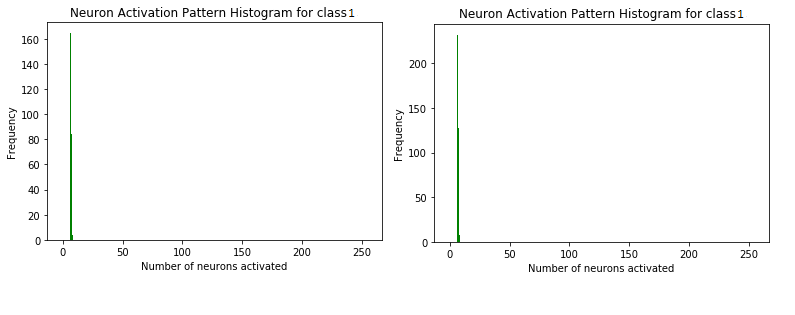}
\caption{NAP for pedestrian. Left is with classification confidence 95\% while right is with 80\%}
\label{fig:NAP_pedestrian}
\end{figure}

\begin{figure}[!htb]
\centering
\includegraphics[width=\linewidth]{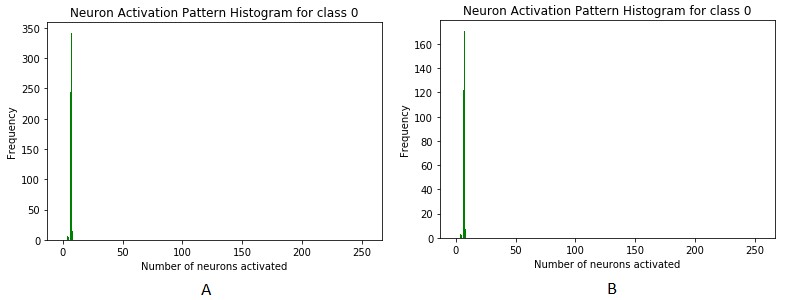}
\caption{NAP for vehicle. A is with classification confidence 95\% while B is with 80\%}
\label{fig:NAP_vehicle}
\end{figure}

NAP results for pedestrian class are shown in Figure~\ref{fig:NAP_pedestrian}, and for vehicle class in Figure~\ref{fig:NAP_vehicle}.

Neuron K-projection coverage was applied on the same layer and for K=2, results are summarized in Table~\ref{tab:k_projection}.

\begin{table}[!htb]
\caption {Results of k-projection coverage with k=2}
 \centering
 \renewcommand{\arraystretch}{1.2}
\begin{tabular}{|c|c|c|} 
\hline
Input size (images)  & Active patterns  & Coverage (\%) \\ 
\hline
40     &  32893/129540  & 25.39 \\ 
\hline
509    &  33149/129540  & 25.58\\ 
\hline
\end{tabular}
\label{tab:k_projection} 
\end{table}

For synthetic noise analysis, four models were evaluated:
\begin{itemize}
 \item \textit{YOLOv3}  was the standard model that was trained on the COCO dataset \cite{coco}.
 \item \textit{Model A} was trained on ISSD's tunnel pedestrian dataset using transfer learning on YOLOv3.
 \item \textit{Model B} was trained on both ISSD's tunnel pedestrian and the synthetic noise dataset which was created with NNDK's noise generation package.
 \item \textit{Model C} was the modified version of model A by further applying transfer learning (10 extra epochs) with the noise dataset to tune its parameters accordingly. 
\end{itemize}

Models performance analysis against synthetic noise using average perturbation loss metric is shown in Table~\ref{tab:perturbation_loss}. Models (A, B, C) are compared with the perturbation loss on the standard YOLOv3. The metric measures the efficacy of synthetic noise. Therefore, the lower the metric, the better the model. 

\begin{table}[!htb]
 \caption {Performance analysis of models against synthetic noise }
 \centering
 \renewcommand{\arraystretch}{1.2}
\begin{tabular}{|c|c|c|c|c|c|c|} 
\hline
\multirow{2}{*}{Model} & \multicolumn{6}{c|}{Average loss (\%)}\\
\cline{2-7}
& Gaussian & Poisson & S\&P & Snow & Haze & Fog \\
\hline
YOLOv3 & 95 & 95 & 95 & 95 & 95 & 95 \\
\hline
A  & 95  & 94.2  & 94.3 & 95 & 95 & 95 \\
\hline
B & 93.8 & 93 & 93.2 & 94 & 94 & 95 \\
\hline
C & 92 & 92.7 & 92.7 & 94.5 & 92.5 & 94 \\
\hline
\end{tabular}
\label{tab:perturbation_loss} 
\end{table}

\end{appendices}


\begin{thebibliography}{00}
\bibitem{yolov4} Alexey Bochkovskiy, Chien-Yao Wang, and H. Liao, ``YOLOv4: Optimal Speed and Accuracy of Object Detection,'' in ArXiv, vol. abs/2004.10934, 2020. 
\bibitem{nndk} Chih{-}Hong Cheng, Chung{-}Hao Huang, and Georg N{\"{u}}hrenberg, ``nn-dependability-kit: Engineering Neural Networks for Safety-Critical Systems,'' in ArXiv, vol. abs/1811.06746, 2018.
\bibitem{yolov3} Redmon, Joseph and Ali Farhadi. ``YOLOv3: An Incremental Improvement,'' ArXiv abs/1804.02767 (2018): n. pag.
\bibitem{coco} T.-Y. Lin, M. Maire, S. Belongie, J. Hays, P. Perona, D. Ramanan, et al., ``Microsoft COCO: Common objects in context,'' In proceedings European Conference on Computer Vision, pp. 740-755, 2014.
\bibitem{dep_metrics} C. Cheng, G. N{\"{u}}hrenberg, C. Huang, H. Ruess and H. Yasuoka, ``Towards Dependability Metrics for Neural Networks,'' 2018 16th ACM/IEEE International Conference on Formal Methods and Models for System Design (MEMOCODE), Beijing, 2018, pp. 1-4, doi: 10.1109/MEMCOD.2018.8556962.
\bibitem{max_res} Cheng CH., N{\"{u}}hrenberg G., Ruess H. ``Maximum Resilience of Artificial Neural Networks,'' In: D'Souza D., Narayan Kumar K. (eds) Automated Technology for Verification and Analysis. ATVA. Lecture Notes in Computer Science, vol 10482. Springer, 2017.
\bibitem{runtime_monitor} Chih{-}Hong Cheng, Georg N{\"{u}}hrenberg, and Hirotoshi Yasuoka. ``Runtime Monitoring Neuron Activation Patterns,'' Design, Automation \& Test in Europe Conference \& Exhibition, 300-303. 10.23919/DATE.2019.8714971, 2019.
\bibitem{combinatorial} C. J. Colbourn, ``Combinatorial aspects of covering arrays,'' Le Matematiche, vol 59(1,2):125–172, 2004.
\bibitem{coverage} J. Lawrence, R. N. Kacker, Y. Lei, D. R. Kuhn, and M. Forbes, ``A survey of binary covering arrays. the electronic journal of combinatorics,'' The Electronic Journal of Combinatorics, vol 18(1):84, 2011.
\bibitem{opencv} Bradski, G., and Kaehler, A., ``OpenCV Library,'' Dr. Dobb's journal of software tools, 3, 2000.
\end{thebibliography}
\end{document}